\documentclass[runningheads]{llncs}
\usepackage{array}
\newcolumntype{P}[1]{>{\centering\arraybackslash}p{#1}}
\usepackage{graphicx}
\usepackage{amsfonts}
\usepackage{hyperref}
\usepackage{booktabs}
\usepackage{multirow}
\usepackage{float}
\usepackage[misc]{ifsym}
\usepackage{graphicx,wrapfig,lipsum}
\usepackage{makecell}
\usepackage{xcolor}
\begin{document}
\title{FALFormer: Feature-aware Landmarks self-attention for Whole-slide Image Classification}
%
%
\author{Doanh C. Bui, Trinh Thi Le Vuong \and Jin Tae Kwak \textsuperscript{(\Letter)}}

%
\authorrunning{Bui et al.}
%
\titlerunning{FALFormer}

\institute{School of Electrical Engineering, Korea University, Seoul, Republic of Korea
\email{jkwak@korea.ac.kr}}
\maketitle
\begin{abstract}

Slide-level classification for whole-slide images (WSIs) has been widely recognized as a crucial problem in digital and computational pathology. Current approaches commonly consider WSIs as a bag of cropped patches and process them via multiple instance learning due to the large number of patches, which cannot fully explore the relationship among patches; in other words, the global information cannot be fully incorporated into decision making. Herein, we propose an efficient and effective slide-level classification model, named as FALFormer, that can process a WSI as a whole so as to fully exploit the relationship among the entire patches and to improve the classification performance. FALFormer is built based upon Transformers and self-attention mechanism. To lessen the computational burden of the original self-attention mechanism and to process the entire patches together in a WSI, FALFormer employs Nystr\"om self-attention which approximates the computation by using a smaller number of tokens or landmarks. For effective learning, FALFormer introduces feature-aware landmarks to enhance the representation power of the landmarks and the quality of the approximation. We systematically evaluate the performance of FALFormer using two public datasets, including CAMELYON16 and TCGA-BRCA. The experimental results demonstrate that FALFormer achieves superior performance on both datasets, outperforming the state-of-the-art methods for the slide-level classification. This suggests that FALFormer can facilitate an accurate and precise analysis of WSIs, potentially leading to improved diagnosis and prognosis on WSIs.

\end{abstract}

\section{Introduction}

In recent years, slide-level whole-slide image (WSI) classification has drawn considerable attention due to its crucial role in clinics for disease diagnosis and prognosis \cite{li2022comprehensive}. Given that WSIs are gigabytes in size, obtaining pixel-level annotations and conducting patch-level classification poses significant challenges to the field of computational pathology. The common strategy to handle and process WSIs is to adopt the multiple instance learning (MIL) paradigm, in which WSIs are divided into a set of disjoint patches, representing WSIs as a bag of patches or instances, and the information from the patches are extracted, selected, and/or aggregated to produce the slide-level prediction. 
There are two main MIL-based approaches including instance-based \cite{hou2016patch,campanella2019clinical} and bag embedding-based models \cite{ilse2018attention,li2021dual,lu2021data,shao2021transmil,zhang2022dtfd}. Instance-based models conduct path-level predictions and then aggregate the results to produce the final prediction for a WSI, while bag embedding-based models map the patches in a bag into one embedding vector and make a prediction based on it. It has recently shown that instance-based models are less efficient than bag embedding-based models \cite{wang2018revisiting,shao2021transmil}. Bag embedding-based models are mostly built based upon an attention mechanism and Transformer architecture. For example, AB-MIL \cite{ilse2018attention} learns to assign a weight for each patch using the attention mechanism, and then performs a weighted average to aggregate all the patch embeddings. Similarly, CLAM \cite{lu2021data} conducts an auxiliary task that clusters the top most-attended patch embeddings as positive patches and the top least-attended patch embeddings as negative patches to constrain and refine the feature space. DTFD-MIL \cite{zhang2022dtfd} is a two-stage MIL-based model, involving sub-MIL and global MIL models, that sought to handle overfitting problems due to the limited number of WSIs. The original patch bag undergoes random splitting to create multiple sub-bags, of which each is processed and aggregated using a sub-MIL model. The representative embeddings from the sub-bags are fed into a global MIL model. {However, long-range dependencies among patches have not been exploited well, as bag embedding is produced by computing a weighted sum of all patches. To address this, some recent studies attempt to make use of self-attention. For instance, TransMIL \cite{shao2021transmil} adopts an architecture of Transformer with positional encoding to retain the spatial information of the cropped patches by reshaping them into 2-D image space and applying multiple learnable convolutions, and utilizes vanilla Nystr\"om Attention to approximate self-attention among patch features. HIGT \cite{higt} introduces a strategy of using clustering and pooling to reduce the number of patches and applies a variant of self-attention. MSPT \cite{msct} performs clustering to reduce the number of patches uses them as queries for self-attention.}


Though successful, these previous MIL-based models, by and large, do not fully explore and utilize all the available patches due to the enormous number of the patches and the computational cost, which likely limits the capability of the model and the subsequent decision making. Hence, advanced methods that can efficiently and effectively process a WSI as a whole or the entire patches together in efficient manner are needed to improve the accuracy and efficiency of the slide-level image classification. 



Herein, we propose a \textbf{F}eature-\textbf{A}ware \textbf{L}andmarks Trans\textbf{Former} (\texttt{FALFormer})
for efficient and effective slide-level image classification. FALFormer is built based upon Transformers and self-attention mechanisms. To reduce the computational burden and to process the entire patches in a WSI, FALFormer adopts  Nystr\"om self-attention \cite{xiong2021nystromformer} which approximates the computation by using a smaller number of tokens or landmarks. For effective learning, FALFormer introduces a \textbf{F}eature-\textbf{A}ware \textbf{L}andmarks Nystr\"om \textbf{S}elf-\textbf{A}ttention (\texttt{FALSA}), which enhances the representation power of the landmarks so as to better approximates the self-attention computation, leading to improved classification performance. Two public datasets, including CAMELYON16 and TCGA-BRCA, are employed to evaluate FALFormer. The experimental results demonstrate that FALFormer is able to conduct the slide-level image classification in an accurate and robust manner and outperforms the state-of-the-art models. Our implementation is available at \footnote{\url{https://github.com/QuIIL/FALFormer}}.


\section{Methodology} \label{sec:method}

\begin{figure*}[http]
\centerline{\includegraphics[width=12.2cm]{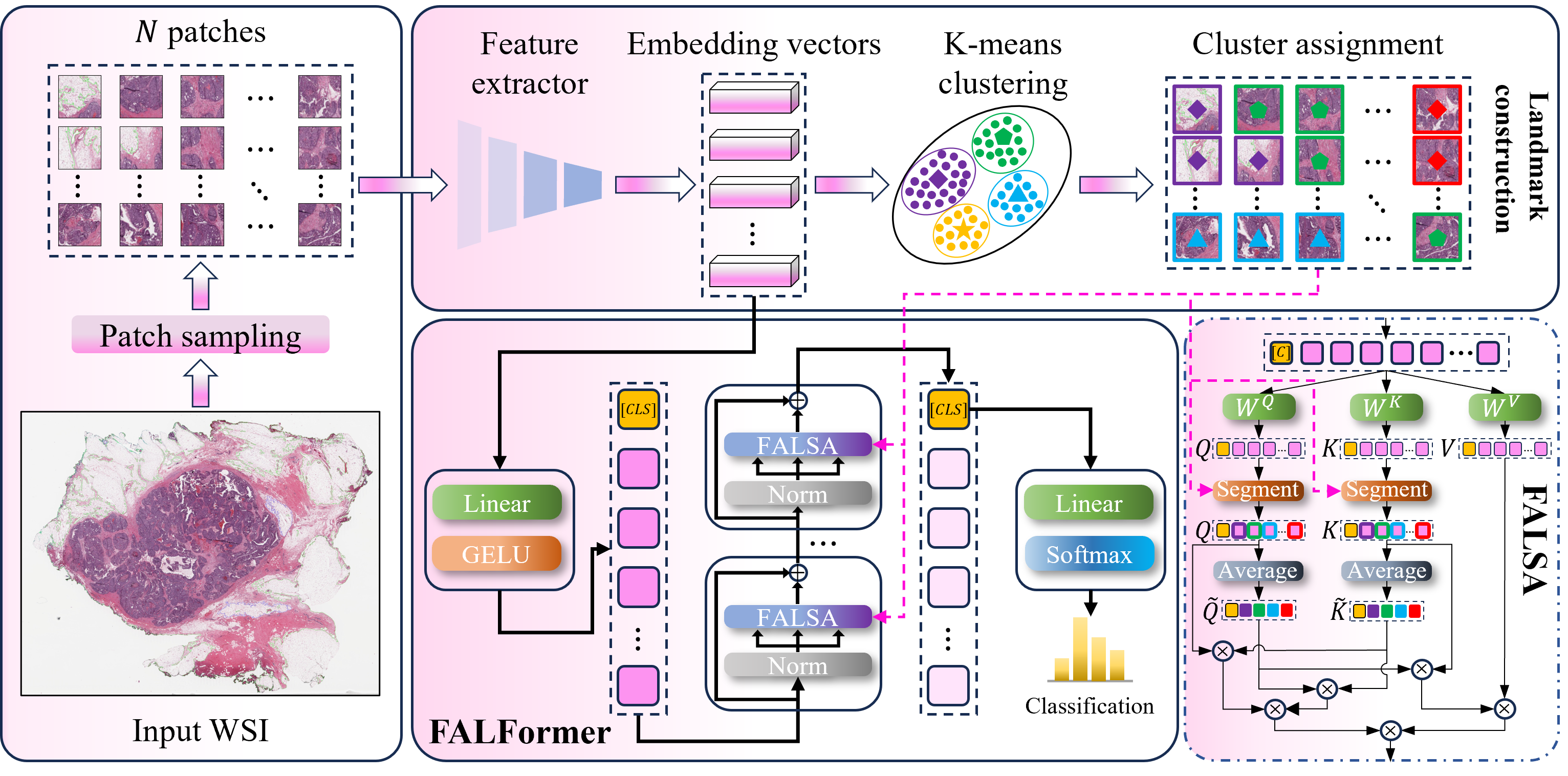}}
\caption{Overview of FALFormer. A WSI is first segmented and tiled into patches based on foreground regions. Then, patch embeddings are extracted and divided into a number of segments by using K-means clustering. FALFormer computes representative landmarks using the segments and use them to process the entire patch embeddings in an efficient and effective mannder for the slide-level classification.}
\label{fig:overview}
\end{figure*}

In this section, we present FALFormer for the WSI classification/sub-typing problem. Let $\mathbf{X}$ be a Giga-sized WSI and $\mathbf{Y}$ be the slide-level class label of $\mathbf{X}$. The objective of our study is to develop a Transformer-based model $\mathcal{T}(\cdot)$, i.e., FALFormer, {which fully exploits entire patches tiled from the WSI with spatially-aware landmarks}, and predict the slide-level label: $\mathbf{Y} = \mathcal{T}(\mathbf{X})$. The overview of FALFormer is illustrated in Fig. \ref{fig:overview}. 


\vspace{-1cm}

\subsection{Feature-aware Landmarks Transformer} \label{sec:network}

Given $\mathbf{X}$, we first remove background to retrieve tissue regions. Subsequently, the tissue regions are tiled into a bag of $N$ patches $\mathbf{P} = \{\mathbf{p}_i\}^{N}_{i=1}$ where $\mathbf{p}_i$ denotes the $i$th patch and $N$ depends on each WSI. From each patch $\mathbf{p}_i$, an embedding vector $\mathbf{h}_i \in \mathbb{R}^{1\times d_f}$, a.k.a. a patch token, is produced using a feature extractor $\mathcal{F}(\cdot)$ where $d_f$ is the dimension of the embedding vector and can vary depending on the choice of $\mathcal{F}(\cdot)$. Finally, a set of patch tokens $\mathbf{H} = \{\mathbf{h}_i\}_{i=1}^{N}$ is obtained.

FALFormer receives $\mathbf{H}$ and conducts a linear projection $\texttt{FC}$ followed by a $\texttt{GELU}$ activation to map the dimension of the patch tokens to the dimension of the model space $d_{model}$. The resultant patch tokens are concatenated with a learnable $\texttt{[CLS]}$ token, denoted as $\mathbf{h}^{(0)}_{\texttt{\small[CLS]}} \in \mathbb{R}^{1 \times d_{model}}$, producing $\mathbf{H}^{(0)} = \texttt{Concatenate}\big(\mathbf{h}^{(0)}_{\texttt{\small[CLS]}}, \texttt{GELU}(\texttt{FC}(\mathbf{H}))\big) \in \mathbb{R}^{(N+1) \times d_{model}}$. 
Then, $\mathbf{H}^{(0)}$ undergoes a stack of $L$ Transformer layers given by: $\mathbf{H}^{(i)} = \texttt{Transformer}^{(i)}\big(\mathbf{H}^{(i-1)}\big), 1 \leq i \leq L$ where $\texttt{Transformer}^{(i)}(\cdot)$ consists of a normalization layer $\texttt{LayerNorm}$ and a $\texttt{FALSA}$. $\texttt{LayerNorm}$ learns affine transform parameters for the patch tokens: $\mathbf{H}' = \texttt{LayerNorm}\big(\mathbf{H}^{(i-1)}\big)$. $\texttt{FALSA}$ is used to approximate the self-attention computation for $\mathbf{H}'$ with a skip connection: $\mathbf{H}^{(i)} = \mathbf{H}^{(i-1)} + \texttt{FALSA}(\mathbf{H}')$. Finally, only the $\texttt{[CLS]}$ token is utilized for the prediction: $\mathbf{Y} = \texttt{Classifier}\Big(\texttt{LayerNorm}\big(\mathbf{h}^{(L)}_{\texttt{\small[CLS]}}\big)\Big)$. 



\subsection{Revisiting Nystr\"om self-attention} \label{sec:nystrom}

Nystr\"om self-attention \cite{xiong2021nystromformer} is an approach to approximate the self-attention computation. Given a sequence of patch tokens $\mathbf{H}$ and three learnable projection matrices $\mathbf{W}^{Q}$, $\mathbf{W}^{K}$, and $\mathbf{W}^{V}$, the standard self-attention computes the query $\mathbf{Q} \in \mathbb{R}^{N \times d_{q}}$, key $\mathbf{K} \in \mathbb{R}^{N \times d_{k}}$, and value $\mathbf{V} \in \mathbb{R}^{N \times d_{v}}$ as follows: $\mathbf{Q} = \mathbf{H}\mathbf{W}^{Q}$, $\mathbf{K} = \mathbf{H}\mathbf{W}^{K}$, $\mathbf{V} = \mathbf{H}\mathbf{W}^{V}$ where $N$ is the number of tokens ($N$ is large) and $d_{q}$, $d_{k}$, and $d_{v}$ is the dimension of the query, key, and value vector, respectively. This requires to compute attention weights $att = \texttt{softmax}\Big(\frac{QK^T}{\sqrt{d_k}}\Big)$, which may cause the out of memory problem due to the large $N$. To lessen the computational burden, Nystr\"om self-attention reduces $N$ to $N'$ ($N' \ll N$) for $\mathbf{Q}$ and $\mathbf{K}$ by grouping $N$ tokens into $N'$ segments where each segment contains $N_k = N/N'$ tokens, producing sets of segments $\{Q^{(i)}\}_{i=1}^{N'}$ and $\{K^{(i)}\}_{i=1}^{N'}$. The tokens are averaged within each segment, forming landmarks $\tilde{\mathbf{Q}} = \big\{\frac{1}{N_k}\Sigma{Q^{(i)}}\big\}_{i=1}^{N'}$ and $\tilde{\mathbf{K}} = \big\{\frac{1}{N_k}\Sigma{K^{(i)}}\big\}_{i=1}^{N'}$. Then, Nystr\"om self-attention can be formulated as:


\begin{equation}
\begin{array}{cc}
\tilde{F} = \texttt{softmax}\Big(\frac{\mathbf{Q}\tilde{\mathbf{K}}^T}{\sqrt{d_q}}\Big), \tilde{A} = \texttt{softmax}\Big(\frac{\tilde{\mathbf{Q}}\tilde{\mathbf{K}}^T}{\sqrt{d_q}}\Big)^{+}, \tilde{B} = \texttt{softmax}\Big(\frac{\tilde{\mathbf{Q}}\mathbf{K}^T}{\sqrt{d_q}}\Big), \\
\tilde{F} \in \mathbb{R}^{N\times N'}, \tilde{A} \in \mathbb{R}^{N'\times N'}, \tilde{B} \in \mathbb{R}^{N'\times N}, \\
\mathbf{H} = (\tilde{F} \times \tilde{A}) \times (\tilde{B} \times \mathbf{V}),
\end{array}
\label{eq:7}
\end{equation}


\noindent where $(\cdot)^+$ denotes the Moore-Penrose pseudoinverse function. The computational complexity of the standard self-attention computation is $\mathcal{O}(N^2)$ because the scale-dot matrix multiplication should be done for all $N$ tokens, which is inappropriate in the case of a large $N$. Nystr\"om self-attention (Eq. \ref{eq:7}) has the computational complexity of $\mathcal{O}(N)$, which is much smaller than $\mathcal{O}(N^2)$ if $N' \ll N$. Though successful, there still remains a question of \textbf{how to choose the landmarks to achieve better performance in the context of WSI?} Nystr\"om self-attention groups the $N$ tokens into $N'$ segments in order from top to bottom. We hypothesize that the better landmarks we choose, the better approximation we obtain, leading to improved classification performance. 

\subsection{Feature-aware Landmarks Nystr\"om Self-Attention (\texttt{FALSA})} \label{sec:falsa}

\texttt{FALSA} chooses the best representative landmarks as follows. First, it defines the maximum number of segments, denoted as $N_s$ ($N_s \ll N$). Second, it utilizes the K-means clustering algorithm to find $N_s$ centroids and to divide the patch tokens $\mathbf{H}$ into $N_s$ segments: $\mathbf{S} = \texttt{K-means}(\mathbf{H}, N_s)$ where $\mathbf{S} = \big\{\mathbf{s}_i\big\}^{N}_{i=1}, s_i \in \{1, 2, ..., N_s\}$, denotes the set of segment IDs for $\mathbf{H}$.
Third, tokens in the query $\mathbf{Q}$ and the key $\mathbf{K}$ are segmented based on their segment IDs $\mathbf{S}$, except for the \texttt{[CLS]} tokens, i.e., $\mathbf{q}_{\texttt{[CLS]}}$ and $\mathbf{k}_{\texttt{[CLS]}}$, to form $\mathcal{Q} = \big\{Q^{(j)}\big\}^{N_s}_{j=1}$, $\mathcal{K} = \big\{K^{(j)}\big\}^{N_s}_{j=1}$. In which, $Q^{(j)}$ and $K^{(j)}$ denote sets of query and key tokens, respectively, belonging to the same $j^{th}$ segment: ${Q}^{(j)} = \big\{\mathbf{q}_i \in \mathbf{Q}|\mathbf{s}_i = j\big\}^{N}_{i=1}$, ${K}^{(j)} = \big\{\mathbf{k}_i \in \mathbf{K}|\mathbf{s}_i = j\big\}^{N}_{i=1}$. 
Fourth, it computes $N_s$ landmarks for the query ($\tilde{\mathcal{Q}}$) and key ($\tilde{\mathcal{K}}$) by computing the average of the tokens within each segment: $\tilde{\mathcal{Q}} = \Big\{\frac{1}{C(j)}\sum Q^{(j)}\Big\}^{N_s}_{j=1}$, $
\tilde{\mathcal{K}} = \Big\{\frac{1}{C(j)}\sum K^{(j)}\Big\}^{N_s}_{j=1}$ where $C(j)$ denotes the the number of the patch tokens belonging to the $j^{th}$ segment. 
Fifth, it concatenates $\tilde{\mathcal{Q}}$ and $\tilde{\mathcal{K}}$ with $\texttt{[CLS]}$ tokens, forming $\mathbf{\tilde{Q}}$ and $\mathbf{\tilde{K}}$: $\mathbf{\tilde{Q}} = \texttt{Concatenate}\big(\mathbf{q}_{\texttt{[CLS]}}, \tilde{\mathcal{Q}}\big), \mathbf{\tilde{K}} = \texttt{Concatenate}\big(\mathbf{k}_{\texttt{[CLS]}}, \tilde{\mathcal{K}}\big)$. Last, $\tilde{\mathbf{{Q}}}$ and $\tilde{\mathbf{{K}}}$ are utilized to compute the Nystr\"om self-attention as described in Eq. \ref{eq:7}.

\section{Experiments and Results} \label{sec:results}

\subsection{Datasets}


\subsubsection{CAMELYON16 \cite{10.1001/jama.2017.14585}.} The dataset was obtained from the CAMELYON16 challenge, which was designed to evaluate algorithms for metastasis detection. {There are 399 WSIs in total, and the official train-test split is used. Specifically, 216, 54, and 129 WSIs are employed for training, validation, and testing, respectively.} WSIs are tiled to 3,617,584 patches, with 9066.6$\pm$6273.6 patches per WSI. In this study, we use CAMELYON16 for tumor vs. non-tumor classification. 

\subsubsection{TCGA-BRCA.} The dataset includes a total of 875 WSIs for breast cancer sub-typing, such as Invasive Ductal Carcinoma (IDC) versus Invasive Lobular Carcinoma (ILC). These annotated WSIs are available on the NIH Genomic Data Commons Data Portal. {Following \cite{chen2022scaling}, we use the ratio of 0.8:0.1:0.1 for the train-val-test split, which are 715 and 79 WSIs for training and validation, respectively, and 81 WSIs for testing.} 2,672,891 patches are generated from the WSIs, with 2567.6$\pm$1592.8 patches per WSI.

\subsection{Implementation Details}

To obtain bag of patches from a WSI, we follow the pipeline provided in the previous study \cite{lu2021data}. For FALFormer, we set the number of transformer layers $L$ to 2 and the model dimension $d_{model}$ to 768, {which are inspired by designs of Vision Transformers and TransMIL.} The number of segments $N_s$ is set to 256. For feature extraction, we utilize two pre-trained models: ResNet-50 \cite{he2016deep}, pre-trained on ImageNet1K, and CTransPath \cite{wang2022transformer}, which is a SwinT-based architecture pre-trained on histopathology datasets. ResNet-50 and CTransPath produce feature vectors of sizes $d_f = 1024$ and $d_f = 768$, respectively. FALFormer is trained for 20 epochs. During training, the RAdam optimizer is utilized, cross-entropy loss is adopted, and the EarlyStopping strategy is employed to halt training if the validation loss does not improve after 10 epochs. {The best model is chosen based on a validation set. The experiment is conducted only once for FALFormer and other models under identical conditions, using the same random seed and environment.}

\subsection{Comparative Study}

For comparison, we include two established slide-level classification models: CLAM \cite{lu2021data} and TransMIL \cite{shao2021transmil}. Both are MIL-based models. CLAM clusters positive and negative patch embeddings within a WSI based on attention scores to improve the bag representation. CLAM can contain a single attention branch (CLAM-SB) and multi-attention branches (CLAM-MB). TransMIL employs a stack of Transformer layers and a positional encoding to capture the spatial information among patch embeddings. We build these three models using the same patch sampling procedure and maintain the hyperparameters from the original works for a fair comparison with FALFormer.

\subsection{Result and Discussions}

We assessed the performance of FALFormer and thee MIL-based models on the two datasets (CAMELYON16 and TCGA-BRCA) using five evaluation metrics including Accuracy (Acc), F1 score (F1), Area under the ROC curve (AUC), Recall, and Precision. 
Table \ref{tab:01} demonstrates the tumor vs. non-tumor classification results on CAMELYON16. Overall, FALFormer with CTransPath achieved the best classification performance of 96.12\% Acc, 0.958 F1, 0.983 AUC, 0.957 Recall, and 0.960 Precision, substantially outperforming other MIL-based models such as $\geq$2.32 Acc\%, $\geq$0.023 F1, $\geq$0.005 AUC, $\geq$0.019 Recall, and $\geq$0.018 Precision. It is worth noting that, using ResNet50 as the feature extractor, there was a consistent performance drop for all the models under consideration. Nonetheless, FALFormer with ResNet50 was superior or comparable to other MIL-based models with ResNet50. These results indicate that the quality of WSI analysis may be dependent on the choice of the feature extractor and the superior performance of FALFormer is not due to a specific choice of the feature extractor.


\begin{table}
\centering
\caption{Results on CAMELYON16 dataset.}
\resizebox{0.8\textwidth}{!}{\begin{tabular}{P{1.8cm}P{2.6cm}P{1.5cm}P{1.5cm}P{1.5cm}P{1.5cm}P{1.5cm}}
\toprule
\textbf{Encoder}  & \textbf{Method} & \textbf{Acc (\%)}   & \textbf{F1}    & \textbf{AUC}   & \textbf{Recall} & \textbf{Precision} \\ \midrule
\multirow{4}{*}{ResNet50}   & CLAM-SB \cite{lu2021data}& \textbf{86.05}  & \textbf{0.849}  & \underline{0.910}  & \textbf{0.844}   & \underline{0.857}      \\
    & CLAM-MB \cite{lu2021data}& \underline{82.95}  & 0.806  & 0.813  & 0.791   & 0.847      \\
    & TransMIL \cite{shao2021transmil} & 82.20      & 0.808      & 0.869      & 0.805       & 0.813  \\
    & FALFormer (ours)      & \textbf{86.05}  & \underline{0.848}  & \textbf{0.934}  & \underline{0.840}   & \textbf{0.860}      \\ \midrule
\multirow{4}{*}{CTransPath} & CLAM-SB \cite{lu2021data} & 88.37  & 0.875  & 0.935  & 0.870   & 0.881      \\
    & CLAM-MB \cite{lu2021data} & \underline{93.80}  & \underline{0.935}  & 0.968  & \underline{0.938}   & 0.931      \\
    & TransMIL \cite{shao2021transmil} & \underline{93.80} & 0.933 & \underline{0.978} & 0.926       & \underline{0.942}  \\
    & FALFormer (ours)      & \textbf{96.12} & \textbf{0.958} & \textbf{0.983} & \textbf{0.957}  & \textbf{0.960} \\
\bottomrule
\end{tabular}}
\label{tab:01}
\end{table}

Table \ref{tab:02} shows the breast cancer sub-typing results on TCGA-BRCA. Similar to the results on CAMELYON16, FALFormer outperformed the three competitors regardless of the choice of the feature extractor, highlighting the strength of FALFormer. FALFormer with CTransPath, in particular, obtained the best classification performance of 96.30\% Acc, 0.937 F1, 0.970 AUC, 0.906 Recall, and 0.978 Precision.
In a head-to-head comparison between ResNet50 and CTransPath, CTransPath always gave a substantial performance gain for FALFormer and other models except Recall for TransMIL; for instance $\geq$3.70\% Acc, $\geq$0.060 F1, $\geq$0.017 AUC, and $\geq$0.059 Precision. 


\begin{table}
\centering
\caption{Results on TCGA-BRCA dataset.}
\resizebox{0.8\textwidth}{!}{\begin{tabular}{P{1.8cm}P{2.6cm}P{1.5cm}P{1.5cm}P{1.5cm}P{1.5cm}P{1.5cm}}
\toprule
\textbf{Encoder}  & \textbf{Method} & \textbf{Acc (\%)}   & \textbf{F1}    & \textbf{AUC}   & \textbf{Recall} & \textbf{Precision} \\ \midrule
\multirow{4}{*}{ResNet50}   & CLAM-SB \cite{lu2021data}& 90.12 & 0.817 & 0.926 & 0.773 & \underline{0.900}      \\
    & CLAM-MB \cite{lu2021data}& \underline{91.36} & \underline{0.844} & \underline{0.942} & 0.805 & \textbf{0.912}      \\
    & TransMIL \cite{shao2021transmil} & 88.89 & 0.787 & 0.932 & \textbf{0.888} & 0.742 \\
    & FALFormer (ours) & \textbf{92.59} & \textbf{0.877} & \textbf{0.945} & \underline{0.860} & 0.899 \\ \midrule
\multirow{4}{*}{CTransPath} & CLAM-SB \cite{lu2021data}& \underline{95.06}  & \underline{0.914} & 0.958 & \underline{0.875}  & \underline{0.971}      \\
    & CLAM-MB \cite{lu2021data}& \underline{95.06}  & \underline{0.914}  & \textbf{0.970} & \underline{0.875}   & \underline{0.971}      \\
    & TransMIL \cite{shao2021transmil}     & 93.83 & 0.888 & 0.949 & 0.844 & 0.964  \\
    & FALFormer (ours) & \textbf{96.30} & \textbf{0.937} & \textbf{0.970} & \textbf{0.906} & \textbf{0.978} \\
\bottomrule
\end{tabular}}
\label{tab:02}
\end{table}

Moreover, we conducted ablation experiments to evaluate the effectiveness of $\texttt{FALSA}$. On CAMELYON16 and TCGA-BRCA, the performance of FALFormer with and without FALSA was measured. FALFormer without FALSA utilizes the original Nystr\"om self-attention as described in Section \ref{sec:nystrom}. The results of the ablation experiments are presented in Table \ref{tab:03}. It is obvious that the adoption of $\texttt{FALSA}$ consistently enhances the classification performance regardless of the dataset and the feature extractor except for Acc, Precision on CAMELYON16 (ResNet50), and AUC, Recall on TCGA-BRCA (CTransPath). This indicates that the quality of landmarks has a direct bearing on the quality of the approximation of the self-attention and the final classification performance. Equipped with stronger landmarks by $\texttt{FALSA}$, FALFormer can facilitate improved analysis of WSIs.

\begin{table}[]
\centering
\caption{Ablation results demonstrating the effectiveness of \texttt{FALSA}.}
\resizebox{1\textwidth}{!}{\begin{tabular}{cP{1.8cm}P{1.5cm}P{1.5cm}P{1.5cm}P{1.5cm}P{1.5cm}|P{1.5cm}P{1.5cm}P{1.5cm}P{1.5cm}P{1.5cm}}
\toprule
\multirow{2}{*}{\textbf{Encoder}} & \multirow{2}{*}{\makecell{\textbf{Self-} \\ \textbf{attention}}} & \multicolumn{5}{c}{CAMELYON16} & \multicolumn{5}{c}{TCGA-BRCA} \\ \cmidrule{3-12} 
 & & \textbf{Acc (\%)} & \textbf{F1} & \textbf{AUC} & \textbf{Recall} & \textbf{Precision}  & \textbf{Acc (\%)}  & \textbf{F1} & \textbf{AUC} & \textbf{Recall} & \textbf{Precision}  \\ \midrule
\multirow{2}{*}{ResNet50}     & Nystr\"om & 86.05 & 0.824 & 0.905 & 0.824 & 0.886 & 90.12 & 0.827 & 0.922 & 0.797 & 0.873 \\
 &\texttt{FALSA} & 86.05 & 0.848 & 0.934 & 0.840 & 0.860 & 92.59 & 0.877 & 0.945 & 0.860 & 0.899  \\ \midrule
\multirow{2}{*}{CTransPath}   & Nystr\"om & 94.57 & 0.941 & 0.948 & 0.933 & 0.953 & 95.06 & 0.926 & 0.971 & 0.946 & 0.909 \\
 & \texttt{FALSA} & 96.12 & 0.958 & 0.983 & 0.957 & 0.960 & 96.30 & 0.937 & 0.970 & 0.906 & 0.978 \\
 \bottomrule
\end{tabular}}
\label{tab:03}
\end{table}

\begin{figure}[H]
\centerline{\includegraphics[width=10cm]{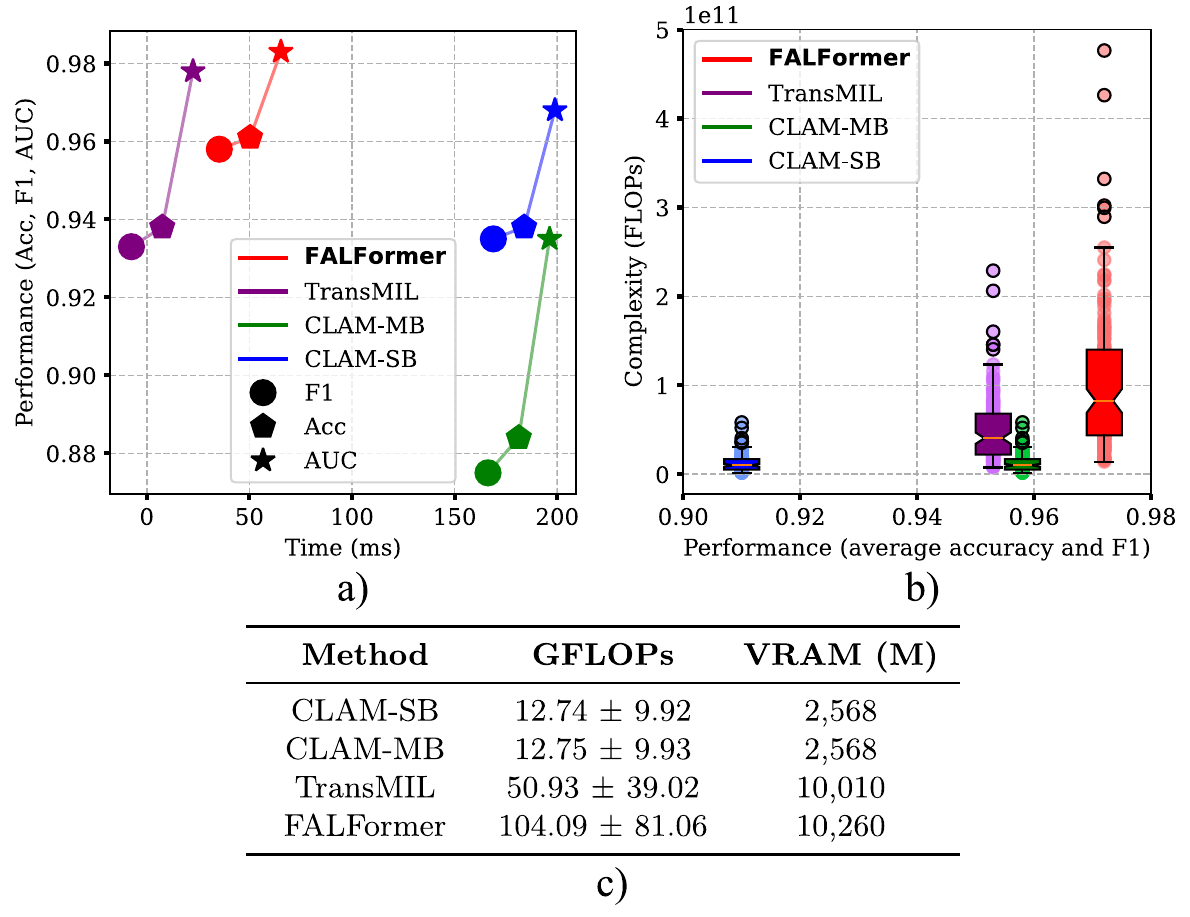}}
\caption{Comparison of Complexity-Performance Trade-off: (a) Performance (Acc, F1, AUC) versus processing time for the most complicated WSI, (b) FLOPs for processing all WSIs versus average Acc and F1, and (c) GFLOPs and VRAM usage.}
\label{fig:complexity}
\end{figure}

We also compared the performance and complexity trade-off within FALFormer and other competitors. We calculated the processing time for a WSI with the highest number of cropped patches, GFLOPs for all WSIs, and VRAM usage on  CAMELYON16 using CTransPath as the feature extractor. These metrics are depicted in Figure \ref{fig:complexity}. FALFormer exhibits the highest GFLOPs and VRAM usage, possibly due to the usage of the entire patches. CLAM models had the smallest GFLOPs and VRAM usage but required the longest processing time. TransMIL demonstrated the shortest processing time and second largest GFLOPs and VRAM usage. The performance of TransMIL was, in general, inconsistent (Table \ref{tab:01} and \ref{tab:02}). Hence, the complexity analysis reveals that FALFormer strikes an acceptable balance between efficiency and accuracy. All measurements are calculated using a single NVIDIA RTX A6000 GPU.

\section{Conclusion}\label{sec:conclusion}

This study introduces FALFormer, a Transformer-based model for efficient and effective WSI classification. FALFormer revisited the Nystr\"om-based self-attention mechanism and proposed \texttt{FALSA}, which leverages the high-level patch features and K-means algorithm to enhance the representative power of the landmarks and the quality of the approximation of the self-attention computation. Equipped with FALSA, FALFormer demonstrates its effectiveness in analyzing WSIs and conducting the slide-level classification.

\section*{Acknowledgments}
This study was supported by a grant of the National Research Foundation of Kroea (NRF) (No. 2021R1A2C2014557 and No. RS-2024-00397293).

\section*{Disclosure of Interests} The authors declare that they have no conflict of interest.
\bibliographystyle{splncs04}
\bibliography{Paper-2447}
\end{document}